\documentclass{article}
\usepackage{spconf,amsmath,graphicx,hyperref}
\usepackage{svg}
\usepackage{cite}
\usepackage{amsmath,amssymb,amsfonts}
\usepackage{algorithmic}
\usepackage{graphicx}
\usepackage{textcomp}
\usepackage{xcolor}
\usepackage{multirow}
\usepackage{pifont}   
\usepackage{amssymb}
\usepackage{array}
\usepackage{fixltx2e}
\usepackage{adjustbox}  
\usepackage{amsmath, graphicx,booktabs,float,multirow,array,makecell,tabularx,url}
\usepackage{url}
\usepackage{caption}
\DeclareCaptionFont{custom}{\fontsize{9pt}{11pt}\selectfont}
\usepackage{subcaption}

\title{Invariant Representation Guided Multimodal Sentiment Decoding with Sequential Variation Regularization}

\name{Guoyang Xu, Zhenxi Song$^{*}$, Junqi Xue, Yuxin Liu, Zirui Wang and Zhiguo Zhang$^{*}$\thanks{$^{*}$Co-corresponding authors: Zhenxi Song (songzhenxi@hit.edu.cn), Zhiguo Zhang (zhiguozhang@hit.edu.cn)}}
\address{Harbin Institute of Technology, Shenzhen, China}
\begin{document}
\ninept
\maketitle
\begin{abstract}
Achieving consistent sentiment representation across diverse modalities remains a key challenge in multimodal sentiment analysis.
However, rapid emotional fluctuations over time often introduce instability, leading to compromised prediction performance.
To address this challenge, we propose a robust sentiment representation \textbf{dual-enhancement strategy} that simultaneously enhances the temporal and modality dimensions, guided by targeted mechanisms in both forward and backward propagation. 
Specifically,
(1) in the modality dimension, we introduce a \textbf{modality-invariant fusion mechanism} that fosters stable cross-modal representations, which aim to capture the common and stable representations shared across different modalities.
(2) in the temporal dimension, we impose a specialized \textbf{sequential variation regularization term} that regulates the model’s learning trajectory during backward propagation, which is essentially total variation regularization degenerated into one-dimensional linear differences.
Extensive experiments on three standard public datasets validate the effectiveness of our proposed approach.
Our code is available at https://github.com/X-G-Y/SATI.
\end{abstract}
%
\begin{keywords}
Multimodal Sentiment Analysis, Representation Disentanglement, Total Variation Regularization, Adversarial Learning.
\end{keywords}
\section{Introduction}
\label{sec:intro}

Multimodal Sentiment Analysis (MSA) has become an active area of research with critical applications across various fields, such as human-computer interaction \cite{hci}, social media analysis \cite{media}, and affective computing \cite{ac}.
The model not only must extract emotional features from different modalities but also needs to capture the consistency of sentiment across the temporal dimension. 
In the process of sentiment recognition, a speaker may exhibit a variety of sentiments. However, the overall sentiment of the video is primarily determined by the dominant sentiment expressed over the extended period. Thus, the stability of emotional fluctuations remains a challenge, especially in long-term sentiment analysis where rapid emotional fluctuations could disturb the model's judgment.

\begin{figure}[t]
    \centering
    \resizebox{0.48\textwidth}{!}
    {\includegraphics{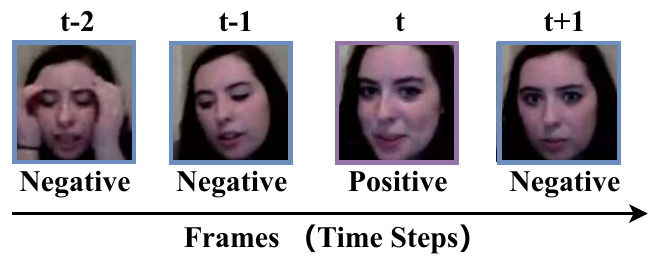}}
    \caption{
    \textbf{Rapid fluctuations in long-term sentiment trends.}
    The frames, shown sequentially from \(t-2\) to \(t+1\), represent discrete time instances in a video. An inconsistent emotional fluctuation disrupts the stable sentiment at frame \(t\).
    }
    \label{emo}
    \vspace{-10pt}
\end{figure}

An example of unstable emotional fluctuation is shown in Fig. \ref{emo}, where the speaker expresses rapid positive sentiment and long-term negative sentiment within the video. The key to accurately recognizing sentiments in such cases lies in designing methods that can effectively distinguish between rapid emotional fluctuations and long-term emotional trends. 

To address this challenge, we enhance the robustness of sentiment representations along both temporal and modality dimensions.
For the temporal dimension, we propose a specific sequential variation regularization, which captures continuous time series patterns within video data by minimizing the distribution between rapid emotional fluctuations and long-term sentiment.
Modality-invariant representations capture the common and stable representations shared across modalities, which are consistent with long-term emotional trends. Therefore, for the modality dimension, we propose an adaptive fusion mechanism that dynamically evaluates the correlations between modalities through invariant features.

\begin{figure*}[htbp]
    \centering
    \resizebox{1\textwidth}{!}
    {\includegraphics{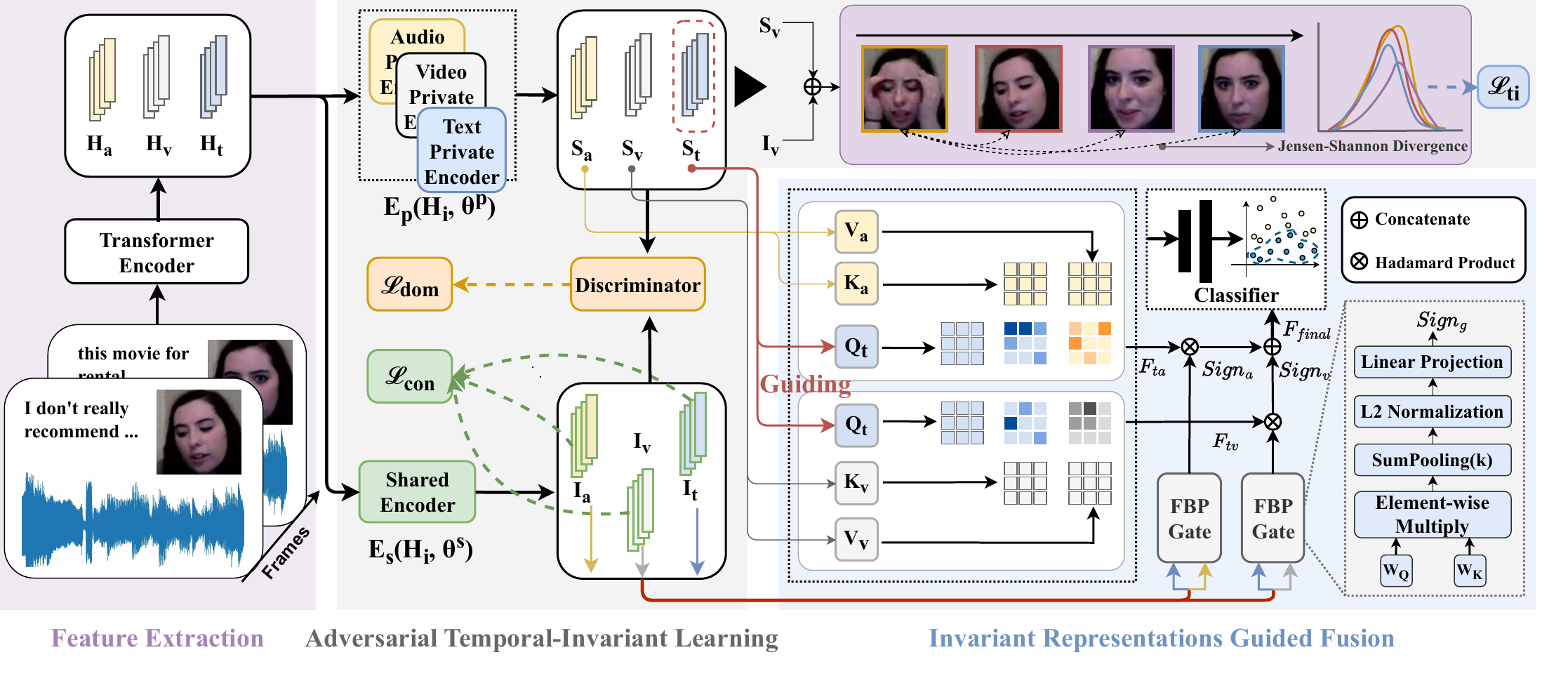}}
    \caption{The overall structure of our proposed model. In the feature extraction module, we begin by enriching the low-level features through the Transformers Encoders to obtain enhanced representations.
    The three processed modality embeddings are fed into shared and private encoders to extract the respective representations subsequently in the representation learning module. 
    Furthermore, the video features are constrained to learn the sequential variation representation.
    Lastly, the modality-specific features are fused in an invariant representation guided manner within the fusion module, gated by the modality-invariant features. The top-right corner of the figure shows that when emotional fluctuations disrupt, the distribution of sentiment evolves accordingly in response to changes in emotional states.}
    \label{model}
\vspace{-3mm}
\end{figure*}

The proposed sequential variation regularization leverages temporal consistency, a property that is crucial in fields such as video compression \cite{compression}, diffusion models \cite{sign}, and image denoising \cite{denoising}. Varghese \textit{et al.} \cite{UTC} improved the continuity of semantic segmentation network predictions over time through temporal consistency. Baltatzis \textit{et al.} \cite{sign} employed temporal consistency to enforce smooth transitions between adjacent frames. However, these methods have not taken into account the potential of temporal consistency in MSA. Compared to the methods mentioned above, our approach differs radically in both methodology and application domains.

The main contributions can be summarized as follows.

(1) We introduce a novel modality fusion approach, which adaptively steers the interactions between different modalities to address modality heterogeneity by modality-invariant representations.


(2) The proposed sequential variation regularization facilitates the identification of holistic structures and relationships within time series, ensuring that the learned representations remain stable and consistent in the presence of rapid emotional fluctuations and noise.

(3) Extensive experiments conducted on three widely used benchmark datasets demonstrate that the proposed approach achieves new state-of-the-art results, and the ablation studies further validate the effectiveness of each component.
\section{PROPOSED MODEL}
\label{sec:PROPOSED MODEL}

The overall architecture of our model is shown in Fig. \ref{model},
consisting of the feature extraction module, representation learning module, 
and invariant representations guided fusion module.

\begin{figure}[t]
    \centering
    \resizebox{0.48\textwidth}{!}
    {\includegraphics{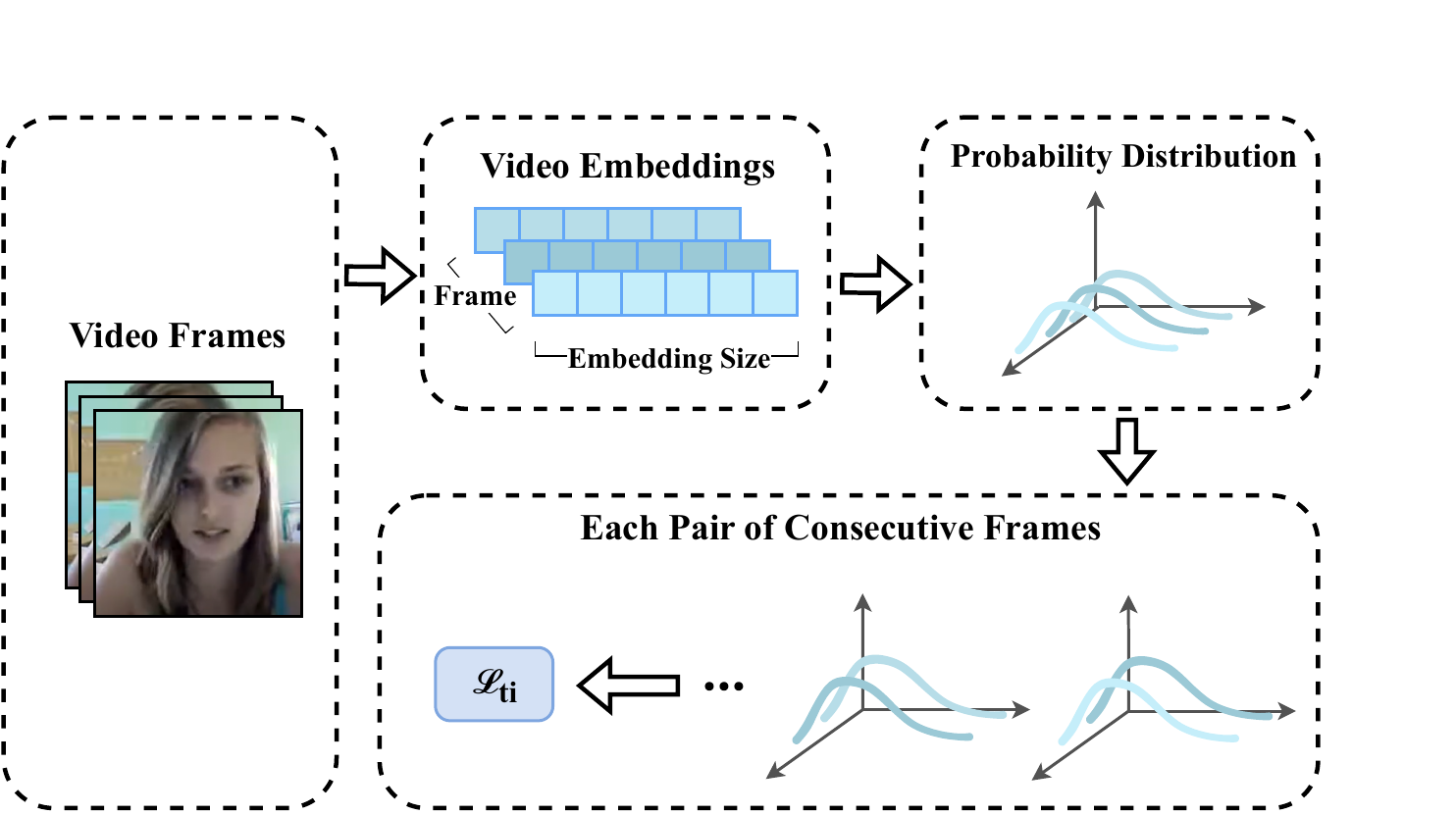}}
    \caption{The details of sequential variation regularization.}
    \label{TILOSS}
    \vspace{-10pt}
\end{figure}

\subsection{Feature Extraction}
For language modality, we feed the input text into RoBERTa \cite{RoBERTa} to extract the text representations. For other modalities, we use one layer Transformer Encoders to capture long-range dependencies.
The outputs of each modality are denoted as $H_i$, where $i \in \{a, v, t\}$.

\subsection{Adversarial Modality Disentanglement}
\textbf{Modality-Invariant and Modality-Specific Representations Learning.}
Our model leverages a shared encoder to capture invariant representations of different modalities, effectively reducing the heterogeneity gap.
To learn the specific representations, our model utilizes three different private encoders that map modality embeddings to the modality-specific subspaces.
The invariant representations $I_i$ and specific representations $S_i$ are denoted as:
\vspace{-2mm}
\begin{equation}
    I_i = E_I(H_i, \theta^I),   S_i = E_S(H_i, \theta^S) 
\vspace{-2mm}
\label{eqencoder}
\end{equation}
where shared encoder $E_I$ shares the parameters $\theta^I$ and private encoders $E_S$ assign separate parameters $\theta^S$ for each modality.

To align the different modalities representations in the invariant subspace, we apply the consistency loss to disentangled representation learning.
We use the Central Moment Discrepancy (CMD) \cite{CMD} to measure the difference between the two modalities.
The consistency loss can be calculated as:
\vspace{-2mm}
\begin{equation}
    \mathcal{L}_\text{con} = \frac{1}{3} \sum_{m_1, m_2 \in \{a, v, t\}} \text{CMD}(I_{m_1}, I_{m_2})
\vspace{-2mm}
\end{equation}
\textbf{Adversarial Learning.}
We introduce a modality discriminator to encourage the shared and private encoders to produce distinct representations. 
The invariant and specific representations are fed into the discriminator  after passing through gradient reversal layers \cite{reversal}, then the discriminator identifies the modality of the representation.:
\begin{equation}
    \mathcal{D}({h_i}, \theta_\mathcal{D}) = \text{softmax}(\boldsymbol{W}_\mathcal{D}^T \cdot \text{Linear}({h_i}))
\end{equation}
where $h_i \in \{I_i, S_i\}$ and $\boldsymbol{W}_\mathcal{D}$ is a learnable parameter matrix.
Following the previous work \cite{MM} , we apply the additive angular margin loss  $\mathcal{L}_{am}$ \cite{aml} to enhance the intra-class compactness and inter-class discrepancy.
The domain loss can be calculated by:
\vspace{-2mm}
\begin{equation}
\mathcal{L}_\text{dom} = \frac{1}{n} \sum_{i=1}^{n} \sum_{m \in \{a, t, v\}} (\mathcal{L}_{am}(I_m, y_m) + \mathcal{L}_{am}(S_m, y_m))
\end{equation}
where $y_m$  denotes the ground-truth modality label.

\begin{table*}[htbp]
    \caption{The Experiment Results on CMU-MOSI and CMU-MOSEI}
    \small
    \begin{center}
    \fontsize{9pt}{11pt}\selectfont 
    {
    \begin{tabular}{lcccccccccc}
    \hline
    \multirow{2}{*}{\textbf{Model}} & \multicolumn{5}{c}{\textbf{CMU-MOSI}} & \multicolumn{5}{c}{\textbf{CMU-MOSEI}} \\
    \cline{2-11} 
     & \textbf{{MAE}}& \textbf{{Corr}}& \textbf{Acc-2} & \textbf{F1-Score} & \textbf{Acc-7} & \textbf{MAE} & \textbf{Corr} & \textbf{Acc-2} & \textbf{F1-Score} & \textbf{Acc-7} \\
    \hline
    MISA\cite{MISA}& 0.783 & 0.761 & {81.8/83.4} & {81.7/83.6} & 42.3 &0.555&0.756&83.6/85.5&83.8/85.3&52.2 \\
    RegBn\cite{RegBn}& - & 0.691 & {81.8/-} & {82.3/-} & 38.6 &-&0.666&81.1/-&81.2/-&50.5\\
    MMIN\cite{MMIN}& 0.741 & 0.795 & {83.53/85.52} & {83.46/85.51} & -&0.542&0.761&83.84/85.88&83.91/85.76&- \\
    TDMER\cite{TDMER}&0.712 & - & {-/86.10} & {-/86.00} & 43.1 & {0.532}&-&- /85.90 &-/85.70& \textbf{54.3} \\
    CAGC\cite{CAGC}& 0.775 & 0.774 & {-/85.70} & {-/85.60} & 44.80  &-&-&-&-&-\\
    Self-MM\cite{SelfMM}& 0.713 & 0.798 & {84.00/85.98} & {84.42/85.95} & -&0.530&0.765&82.81/85.17&82.53/85.30&-\\
    FDMER\cite{MM}& 0.724 & 0.788 & {84.6/-} & {84.7/-} &44.1&0.536&0.773&86.1/-&85.8/-&54.1\\
    \textbf{Ours} &  \textbf{0.683} &  \textbf{0.814} &  \textbf{85.13/86.89} &  \textbf{85.08/86.90} &  \textbf{45.63} & \textbf{0.528}& \textbf{0.795}& \textbf{86.12/86.55}& \textbf{85.97/86.21}& {52.56}\\
    \hline
    \multicolumn{4}{l}{$^{\mathrm{a}}$ The best results are labeled in bold.}
    \end{tabular}}
    \label{mosi_mosei}
    \end{center}
    \vspace{-8mm}
    
\end{table*}

\subsection{Sequential Variation Regularization.}
Our regularization strategy aims to maintain feature stability across time steps, ensuring that the learned representations are resilient to temporal variations. To mitigate the impact of emotional fluctuations, our optimization objective is defined as minimizing the distance between the feature distributions of frames exhibiting emotional fluctuations and representing long-term emotional trends:
\vspace{-2mm}
\begin{equation}
\min  \left\| {{f}_{fluct}} - {f_{domin}} \right\|^2 
\end{equation}
where ${f}_\text{fluct}$ represents the features of frames in emotional fluctuations, and ${f_\text{domin}}$ represents the features of frames corresponding to the dominant sentiment in the video.



Considering that significant changes only disrupt when there is a notable shift in sentiment, we simplify the optimization goal to minimize the distance between the feature distributions of adjacent frames around the fluctuating emotional period, and further extend this to minimize the distance between the feature distributions of any adjacent frames. 
We further incorporate a TV regularization term to promote temporal smoothness:
\vspace{-2mm}
\begin{equation}
\text{TV}_2(f) = \int_\text{fluct+domin} \|\nabla f(x)\|^2 \, dx
\vspace{-2mm}
\end{equation}
To reduce the computational overhead of second-order differences, we degenerate the TV regularization into its one-dimensional form and employ the Jensen–Shannon Divergence (JSD) \cite{JS} to quantify the differences between feature distributions.
Thus, our proposed specialized sequential variation regularization, as illustrated in Fig. \ref{TILOSS}, can be therefore implemented via a temporal invariant loss $\mathcal{L}_\text{ti}$ as:
\vspace{-2mm}
\begin{equation}
    \mathcal{L}_\text{ti} = \frac{1}{n-1} \sum_{i = 1}^{n-1} \text{JSD}(\text{softmax}(R_{i}), \text{softmax}(R_{i+1}))
\end{equation}
where $n$ represents the number of time steps in the video data and $R_{i}$  represents the video representations at the $i$-th time steps.

\subsection{Invariant Representation Guided Fusion Module}

\textbf{Fusion Procedure.}
Sentiments in video, audio, and text often exhibit transient, intense variations, such as sudden changes in tone or abrupt facial expressions. These rapid fluctuations usually reflect noise rather than the underlying long-term emotional trend. Directly fusing raw modality-specific features may cause the model to be biased by these transient anomalies, leading to unstable predictions.

To address this issue, we use modality-invariant representations to guide the interactions, which aim to capture the common and stable representations shared across different modalities different from the previous work \cite{if}. Specially, invariant representation guided fusion module has two parallel inter-modality attention streams with respective gate-controlled mechanisms.
To enhance modality alignment, both streams are driven by the text modality to provide consistent context.

After positional encoding, the modality-specific features $S_i$ ($i \in \{a, v\}$) and $S_t$ are processed through the cross-attention stream  to produce the interacted features $F_{ti}$ ($i \in \{a, v\}$):
\vspace{-2mm}
\begin{equation}
    F_{ti} = Attention(S_t, S_i, S_i) = softmax\left(\frac{S_t S_i^T}{\sqrt{d_k}}\right)S_i
\label{eqattn}
\vspace{-2mm}
\end{equation}
Meanwhile, the gated mechanism takes modality-invariant features $I_i$ ($i \in \{a, v\}$) and $I_t$ as inputs, producing strong correlations between modality-invariant features at each time step to control the interactions of modality-specific features.

During modality representation learning, the modality-specific features develop more distinct representations, making it difficult to accurately measure inter-modality similarity. By contrast, the invariant features capture the shared information across modalities, providing a more reliable basis for assessing correlation.


Our gated mechanism employs the Factorized Bilinear Pooling (FBP) \cite{FBP} to generate the temporal gated signals $Sign_g$ ($g \in \{a, v\}$), as illustrated in Fig. \ref{model}.
The final representation $F_{final}$ is defined as:
\vspace{-2mm}
\begin{equation}
    F_\text{final} = \text{concatenate}(Sign_a \cdot F_{ta}, Sign_v \cdot F_{tv})
\label{eqmul}
\vspace{-2mm}
\end{equation}

\textbf{Prediction.}
We feed the fused representation into an MLP to obtain the prediction output. 
The final loss function is expressed as follows:
\begin{equation}
    \mathcal{L} = \mathcal{L}_\text{task} + \alpha \mathcal{L}_\text{con} + \beta \mathcal{L}_\text{dom} + \gamma \mathcal{L}_\text{ti}
\end{equation}
where \(\alpha\), \(\beta\), and \(\gamma\) are the trade-off parameters and $L_\text{task} \in \{L_\text{MSE}, L_\text{CE}\}$ stands for the loss prediction function for different tasks.

\section{EXPERIMENTS}
\subsection{Datasets}
We evaluate our approach on two widely used multimodal sentiment analysis datasets-CMU-MOSI \cite{mosi}, CMU-MOSEI \cite{mosei}, and a humor analysis dataset-UR\_FUNNY \cite{urfunny}.

\subsection{Evaluation Criteria}
Following the previous works \cite{MISA, SelfMM, TDMER}, we utilize five evaluation metrics to assess the performance of the proposed model on CMU-MOSI, CMU-MOSEI and UR\_FUNNY. 
Specifically, we report binary classification accuracy (Acc-2), seven-class classification accuracy (Acc-7), and weighted F1 score (F1-Score) for the classification task as well as mean absolute error (MAE) and Pearson correlation (Corr) for the regression task. For Acc-2 and F1-Score, we use the segmentation marker -/- to report the results, with the left score representing "negative/non-negative" classification and the right score representing "negative/positive" classification.
For UR\_FUNNY, we report binary classification accuracy (Acc-2).

\subsection{Parameters Settings.}
In our experiments, we utilize the segmentation methods offered by the CMU-Multimodal SDK \cite{multiSDK}.
For Roberta and LSTM, we employed the last hidden state as output. 
For the CMU-MOSI, CMU-MOSEI, and UR\_FUNNY datasets, the learning rate of the model is 2e-5, 5e-5, and 2.5e-5, respectively. During the training process on the UR\_FUNNY dataset, we froze the parameters of the RoBERTa. 
Due to memory limitations and convergence speed considerations, we set the batch sizes for CMU-MOSI, CMU-MOSEI, and UR\_FUNNY to 64, 16, and 128, respectively.
Furthermore, we standardize \(\alpha\) as 1.0, \(\beta\) as 0.4, and \(\gamma\) as 1 for all experiments.

\begin{table}[t]
    \caption{The Experiment Results on UR\_FUNNY}
    \centering
    \fontsize{9pt}{11pt}\selectfont 
        \begin{tabular}{lccc}
        \hline
        \textbf{Model} & \textbf{Context}& \textbf{Target}& \textbf{Acc-2} \\
        \hline
        C-MFN\cite{urfunny}  &\checkmark& & 58.45\\
        C-MFN\cite{urfunny}  & &\checkmark & 64.47 \\
        TFN\cite{TFN}   & &\checkmark & 64.71 \\
        C-MFN\cite{urfunny} &\checkmark & \checkmark& 65.23 \\
        MISA* \cite{MISA}  & &\checkmark & 68.41 \\
        \textbf{Ours} & &\checkmark & \textbf{70.32} \\
        \hline
        \multicolumn{4}{l}{$^{\mathrm{a}}$MISA with $\ast$ are reproduced under the same conditions.}
        \end{tabular}
    \label{ur_funny}
    \vspace{-3mm}
\end{table}

\begin{table}[t]
    \caption{Ablation Study on CMU-MOSI}
    \vspace{-5mm}
    \begin{center}
    \fontsize{9pt}{11pt}\selectfont 
    \resizebox{0.5\textwidth}{!}{
    \begin{tabular}{lccccc}
    \hline
    \textbf{Strategies}& \textbf{Acc-2} &\textbf{F1} &\textbf{MAE} & \textbf{Corr} & \textbf{Acc-7} \\
    \cline{1-6} 
    \textbf{Ours}&\textbf{85.13/86.89} & \textbf{85.08/86.90} & \textbf{0.683} & \textbf{0.814} & \textbf{45.63} \\
    w/o AL& 83.09/84.76 & 83.06/84.78 & 0.716 & 0.792 & 45.19\\
    w/o SVR& 83.82/85.06 & 83.86/85.14 & 0.737 & 0.797 & 43.29 \\
    w/o IRGM & 82.80/84.30 & 82.82/84.37 & 0.729 & 0.789 & 45.04 \\
    \hline
\end{tabular}
    }
    \label{Ablation}
    \end{center}
    \vspace{-8mm}
\end{table}

\begin{table}[t]
    \caption{Noise Robustness Study on CMU-MOSI}
    \begin{center}
    \fontsize{9pt}{11pt}\selectfont 
    \begin{tabular}{lc|cc}
    \hline
    \textbf{Strategies}& \textbf{Noise} & \textbf{Acc-2} &\textbf{F1}  \\
    \cline{1-4} 
    \multirow{4}{*}{Ours}& \ding{55} & \textbf{85.13/86.89} & \textbf{85.07/86.90} \\
    &(0, 0.1)  & 84.84/86.59 & 84.79/86.59\\
    &(0, 0.5)  & 83.97/85.82 & 83.89/85.84  \\
    &(0, 1.0)  & 84.26/85.98 & 84.18/86.01   \\
    \cline{1-4} 
    \multirow{4}{*}{w/o SVR}& \ding{55} & 83.38/85.37 & 83.39/85.31   \\
    & (0, 0.1) & 82.09/84.21 & 82.05/84.19  \\
    & (0, 0.5) & 81.94/83.76 & 81.92/83.72  \\
    & (0, 1.0) & 81.80/83.60 & 81.81/83.54  \\
    \hline
    \end{tabular}
    \label{Robustness}
    \end{center}
\end{table}

\subsection{Comparison with Baselines}
To evaluate the rationality and effectiveness of our method, we compare the proposed model with the following recent and competitive baselines: 
MISA \cite{MISA}, RegBn \cite{RegBn}, MMIN \cite{MMIN}, TDMER \cite{TDMER}, CAGC \cite{CAGC}, Self-MM \cite{SelfMM}, FDMER \cite{MM}, MFN \cite{urfunny}, and TFN \cite{TFN}.

\textbf{Multimodal Emotion Recognition.}
The results compared with baselines on the two datasets are presented in TABLE \ref{mosi_mosei}, leading to the following observations.
Our method achieves significant improvements over previous state-of-the-art approaches across all metrics on both benchmarks, except the seven-class classification task and MAE on the CMU-MOSEI dataset.
The performance degradation in the seven-class classification task and MAE arises from the model's overemphasis on adversarial learning at the expense of inter-class classification.
Compared to MISA \cite{MISA}, which also learns different subspace representations, our method shows that an adversarial approach more effectively disentangles modality-invariant and modality-specific subspaces. 
Similarly, compared to the recent MMIN \cite{MMIN}, which captures the modality-specific characteristics at a coarse-grained level, our invariant representation guided fusion approach achieves effective multimodal representation learning with a simpler structure.

\textbf{Multimodal Humor Recognition.}
To validate the generalizability of our model, we conducted comparative experiments on the UR\_FUNNY dataset. Table \ref{ur_funny} shows that our model performs well in learning effective representations of heterogeneous modalities.

\subsection{Ablation Studies}
We conducted ablation studies to evaluate the impact of individual components on overall performance, including Adversarial Learning (AL), Sequential Variation Regularization (SVR) and Invariant Representation Gated Mechanism (IRGM). As shown in TABLE \ref{Ablation}, the results highlight the importance of each module.

To evaluate the robustness of the model to noise, we introduced Gaussian noise into the initially extracted features, with varying distributions \(N(0, 0.1)\), \(N(0, 0.5)\), and \(N(0, 1.0)\). 
As shown in TABLE \ref{Robustness}, the results indicate that the addition of noise has minimal impact on the overall model performance. 


Compared to the model without the sequential variation regularization, our model exhibits significantly less performance degradation under noisy conditions, demonstrating the effectiveness of the sequential variation regularization in enhancing robustness. Furthermore, it is noteworthy that our model achieves better robustness when Gaussian noise with a distribution of \(N(0, 1.0)\) is added, compared to \(N(0, 0.5)\). This phenomenon may be attributed to a potential regularization effect introduced by noise with a standard deviation of 1.0, which forces the model to rely less on fine-grained details and focus on more robust, global patterns in the data.

\begin{figure}[t]
    \centering
    \vspace{-5mm}
    \resizebox{0.48\textwidth}{!}
    {\includegraphics{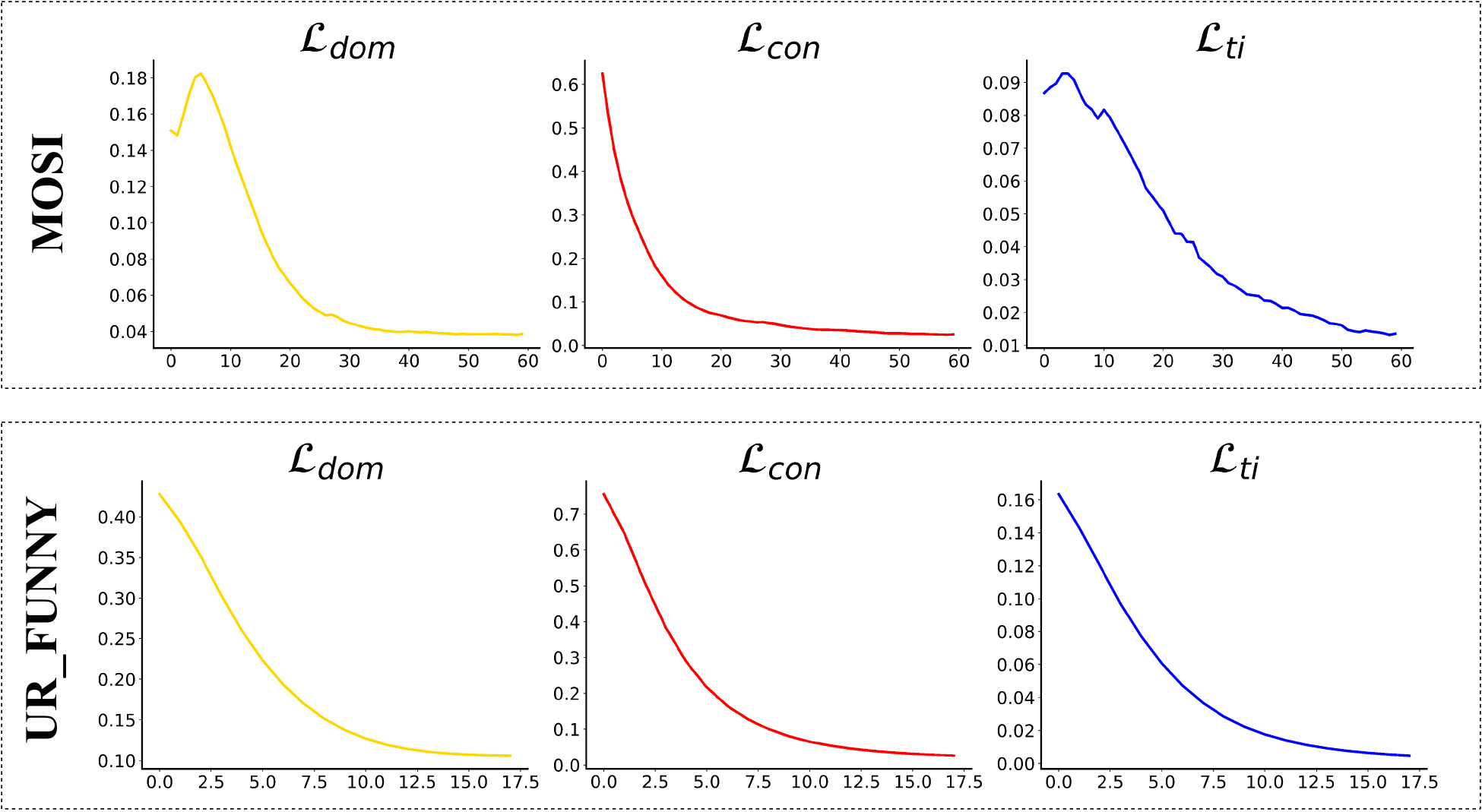}}
    \caption{Training visualization of the ${\mathcal{L}_\text{dom}}$, ${\mathcal{L}_\text{con}}$ and ${\mathcal{L}_\text{ti}}$.}
    \label{loss}
    \vspace{-5mm}
\end{figure}

\subsection{Visualization Results}
The loss metrics serve as key indicators for evaluating the model's capacity to learn diverse representations. Observations from Fig. \ref{loss} reveal that all loss metrics consistently decrease as the number of epochs increases on the MOSI dataset, indicating that the model is effectively learning representations as intended. Similar trends were observed on the UR\_FUNNY dataset, further validating the model's ability to generalize across different tasks and datasets.

\section{CONCLUSION}


In this paper, we address the challenge of robust and consistent multimodal sentiment analysis through a two-dimensional perspective: modality learning and temporal consistency. 
At the modality level, we propose an invariant representations guided fusion module to capture stable cross-modal information and mitigate the impact of short-term emotional fluctuations.
At the temporal level, we introduce sequential variation regularization to ensure the stability of multimodal representations over time, while providing interpretability in multimodal sentiment analysis.
Experimental results validate the superiority of our approach, achieving state-of-the-art performance across multiple benchmarks, including CMU-MOSI, CMU-MOSEI, and UR\_FUNNY datasets.

\vfill\pagebreak
\sloppy
\section{acknowledgements}
{\fontsize{9}{11}\selectfont
This work partially was supported by the Shenzhen Science and Technology Program ( Grant Nos. RCBS20231211090800003, ZDSYS20230626091203008 ).
}
\bibliographystyle{IEEEbib}
\bibliography{strings,refs}

@Article{hci,
  author    = "Moin, A. and Aadil, F. and Ali, Z. and others",
  title     = "Emotion recognition framework using multiple modalities for an effective human–computer interaction",
  journal   = "The Journal of Supercomputing",
  year      = "2023",
  volume    = "79",
  number    = "8",
  pages     = "9320-9349"
}

@InProceedings{media,
  author       = "Morency, L. P. and Mihalcea, R. and Doshi, P.",
  title        = "Towards multimodal sentiment analysis: Harvesting opinions from the web",
  booktitle    = "Proceedings of the 13th International Conference on Multimodal Interfaces",
  organization = "ACM",
  year         = "2011",
  pages        = "169-176"
}

@Article{ac,
  author    = "Wang, Y. and Song, W. and Tao, W. and others",
  title     = "A systematic review on affective computing: Emotion models, databases, and recent advances",
  journal   = "Information Fusion",
  year      = "2022",
  volume    = "83",
  pages     = "19-52"
}

@InProceedings{compression,
  author       = "Pessoa, J. and Aidos, H. and Tomás, P. and others",
  title        = "End-to-end learning of video compression using spatio-temporal autoencoders",
  booktitle    = "Proceedings of the IEEE Workshop on Signal Processing Systems",
  organization = "IEEE",
  year         = "2020",
  pages        = "1-6"
}

@InProceedings{sign,
  author       = "Baltatzis, V. and Potamias, R. A. and Ververas, E. and others",
  title        = "Neural Sign Actors: A diffusion model for 3D sign language production from text",
  booktitle    = "Proceedings of the IEEE/CVF Conference on Computer Vision and Pattern Recognition",
  organization = "IEEE",
  year         = "2024",
  pages        = "1985-1995"
}

@InProceedings{denoising,
  author       = "Zhou, Y. and Xu, X. and Shen, F. and others",
  title        = "Temporal denoising mask synthesis network for learning blind video temporal consistency",
  booktitle    = "Proceedings of the 28th ACM International Conference on Multimedia",
  organization = "ACM",
  year         = "2020",
  pages        = "475-483"
}

@InProceedings{UTC,
  author       = "Varghese, S. and Gujamagadi, S. and Klingner, M. and others",
  title        = "An unsupervised temporal consistency (TC) loss to improve the performance of semantic segmentation networks",
  booktitle    = "Proceedings of the IEEE/CVF Conference on Computer Vision and Pattern Recognition",
  organization = "IEEE",
  year         = "2021",
  pages        = "12-20"
}

@Article{RoBERTa,
  author    = "Liu, Y. and Ott, M. and Goyal, N. and others",
  title     = "RoBERTa: A Robustly Optimized BERT Pretraining Approach",
  journal   = "arXiv preprint arXiv:1907.11692",
  year      = "2019"
}

@Article{CMD,
  author    = "Zellinger, W. and Grubinger, T. and Lughofer, E. and others",
  title     = "Central moment discrepancy (CMD) for domain-invariant representation learning",
  journal   = "arXiv preprint arXiv:1702.08811",
  year      = "2017"
}

@InProceedings{MISA,
  author       = "Hazarika, D. and Zimmermann, R. and Poria, S.",
  title        = "MISA: Modality-invariant and specific representations for multimodal sentiment analysis",
  booktitle    = "Proceedings of the 28th ACM International Conference on Multimedia",
  organization = "ACM",
  year         = "2020",
  pages        = "1122-1131"
}

@InProceedings{reversal,
  author       = "Ganin, Y. and Lempitsky, V.",
  title        = "Unsupervised domain adaptation by backpropagation",
  booktitle    = "Proceedings of the International Conference on Machine Learning",
  organization = "PMLR",
  year         = "2015",
  volume       = "37",
  pages        = "1180-1189"
}

@InProceedings{aml,
  author       = "Deng, J. and Guo, J. and Xue, N. and others",
  title        = "ArcFace: Additive angular margin loss for deep face recognition",
  booktitle    = "Proceedings of the IEEE/CVF Conference on Computer Vision and Pattern Recognition",
  organization = "IEEE",
  year         = "2019",
  volume       = "44",
  pages        = "4690-4699"
}

@Article{JS,
  author    = "Lin, J.",
  title     = "Divergence measures based on the Shannon entropy",
  journal   = "IEEE Transactions on Information Theory",
  year      = "1991",
  volume    = "37",
  number    = "1",
  pages     = "145-151"
}

@Article{if,
  author    = "Sun, H. and Liu, J. and Chen, Y. W. and others",
  title     = "Modality-invariant temporal representation learning for multimodal sentiment classification",
  journal   = "Information Fusion",
  year      = "2023",
  volume    = "91",
  pages     = "504-514"
}

@InProceedings{FBP,
  author       = "Yu, Z. and Yu, J. and Fan, J. and others",
  title        = "Multi-modal factorized bilinear pooling with co-attention learning for visual question answering",
  booktitle    = "Proceedings of the IEEE International Conference on Computer Vision",
  organization = "IEEE",
  year         = "2017",
  pages        = "1821-1830"
}

@Article{mosi,
  author    = "Zadeh, A. and Zellers, R. and Pincus, E. and Morency, L. P.",
  title     = "Multimodal sentiment intensity analysis in videos: Facial gestures and verbal messages",
  journal   = "IEEE Intelligent Systems",
  year      = "2016",
  volume    = "31",
  number    = "6",
  pages     = "82-88"
}

@InProceedings{mosei,
  author       = "Zadeh, A. B. and Liang, P. P. and Poria, S. and Cambria, E. and Morency, L. P.",
  title        = "Multimodal language analysis in the wild: CMU-MOSEI dataset and interpretable dynamic fusion graph",
  booktitle    = "Proceedings of the 56th Annual Meeting of the Association for Computational Linguistics",
  organization = "ACL",
  year         = "2018",
  volume       = "1",
  pages        = "2236-2246"
}

@Article{urfunny,
  author    = "Hasan, M. K. and Rahman, W. and Zadeh, A. and others",
  title     = "UR-FUNNY: A multimodal language dataset for understanding humor",
  journal   = "arXiv preprint arXiv:1904.06618",
  year      = "2019"
}

@InProceedings{multiSDK,
  author       = "Zadeh, A. and Liang, P. P. and Poria, S. and others",
  title        = "Multi-attention recurrent network for human communication comprehension",
  booktitle    = "Proceedings of the AAAI Conference on Artificial Intelligence",
  organization = "AAAI",
  year         = "2018",
  pages        = "5642-5649"
}

@InProceedings{RegBn,
  author       = "Boozandani, M. G. and Wachinger, C.",
  title        = "RegBN: Batch normalization of multimodal data with regularization",
  booktitle    = "Advances in Neural Information Processing Systems",
  organization = "NeurIPS",
  year         = "2024"
}

@InProceedings{MMIN,
  author       = "Fang, L. and Liu, G. and Zhang, R.",
  title        = "Multi-grained multimodal interaction network for sentiment analysis",
  booktitle    = "ICASSP 2024 IEEE International Conference on Acoustics, Speech and Signal Processing (ICASSP)",
  organization = "IEEE",
  year         = "2024",
  pages        = "7730-7734"
}

@InProceedings{TDMER,
  author       = "Qian, X. and Yu, G. and Chen, L. and others",
  title        = "TDMER: A Task-Driven Method for Multimodal Emotion Recognition",
  booktitle    = "Proceedings of the International Conference on Acoustics, Speech, and Signal Processing",
  organization = "IEEE",
  year         = "2025",
  pages        = "1-5"
}

@InProceedings{CAGC,
  author       = "Sun, K. and Xie, Z. and Ye, M. and others",
  title        = "Contextual augmented global contrast for multimodal intent recognition",
  booktitle    = "Proceedings of the IEEE/CVF Conference on Computer Vision and Pattern Recognition",
  organization = "IEEE",
  year         = "2024",
  pages        = "26963-26973"
}

@InProceedings{SelfMM,
  author       = "Yu, W. and Xu, H. and Yuan, Z. and others",
  title        = "Learning modality-specific representations with self-supervised multi-task learning for multimodal sentiment analysis",
  booktitle    = "Proceedings of the AAAI Conference on Artificial Intelligence",
  organization = "AAAI",
  year         = "2021",
  volume       = "35",
  number       = "12",
  pages        = "10790-10797"
}

@InProceedings{MM,
  author       = "Yang, D. and Huang, S. and Kuang, H. and others",
  title        = "Disentangled representation learning for multimodal emotion recognition",
  booktitle    = "Proceedings of the 30th ACM International Conference on Multimedia",
  organization = "ACM",
  year         = "2022",
  pages        = "1642-1651"
}

@Article{TFN,
  author    = "Zadeh, A. and Chen, M. and Poria, S. and Cambria, E. and Morency, L. P.",
  title     = "Tensor fusion network for multimodal sentiment analysis",
  journal   = "arXiv preprint arXiv:1707.07250",
  year      = "2017"
}

\end{document}